\PassOptionsToPackage{dvipsnames}{xcolor}

\documentclass{article}
\usepackage[utf8]{inputenc} %
\usepackage[T1]{fontenc}    %
\usepackage{hyperref}       %
\usepackage{url}            %
\usepackage{booktabs}       %
\usepackage{amsfonts}       %
\usepackage{nicefrac}       %
\usepackage{microtype}      %
\usepackage{lipsum}		%
\usepackage{graphicx}
\usepackage{float}
\usepackage{natbib}
\usepackage{doi}
\usepackage{xcolor}
\usepackage{lineno}
\usepackage{amsmath}
\usepackage{xcolor}
\usepackage{arxiv}
\usepackage{url}
\newcommand\myshade{85}
\colorlet{mylinkcolor}{violet}
\colorlet{mycitecolor}{blue}
\colorlet{myurlcolor}{Thistle}

\hypersetup{
  linkcolor  = mylinkcolor!\myshade!black,
  citecolor  = mycitecolor!\myshade!black,
  urlcolor   = myurlcolor!\myshade!black,
  colorlinks = true,
}

\usepackage{times}
\usepackage{latexsym}

\usepackage[T1]{fontenc}

\usepackage[utf8]{inputenc}

\usepackage{microtype}

\usepackage{booktabs}
\usepackage{graphicx}
\usepackage{makecell}
\usepackage{multirow}
\usepackage{longtable}
\usepackage{CJKutf8}
\usepackage{float}
\usepackage{algorithm}
\usepackage{algorithmicx}
\usepackage{algpseudocode}

\usepackage{tikz}
\usetikzlibrary{shapes.geometric}

\title{Knowledge Solver: Teaching LLMs to Search for Domain Knowledge from Knowledge Graphs}

\author{Chao Feng, Xinyu Zhang, Zichu Fei}

\begin{document}

\maketitle

\begin{abstract}
Large language models (LLMs), such as ChatGPT and GPT-4, are versatile and can solve different tasks due to their emergent ability and generalizability. However, LLMs sometimes lack domain-specific knowledge to perform tasks, which would also cause hallucination during inference. In some previous works, additional modules like graph neural networks (GNNs) are trained on retrieved knowledge from external knowledge bases, aiming to mitigate the problem of lacking domain-specific knowledge. However, incorporating additional modules: 1) would need retraining additional modules when encountering novel domains; 2) would become a bottleneck since LLMs' strong abilities are not fully utilized for retrieval. In this paper, we propose a paradigm, termed Knowledge Solver (KSL), to teach LLMs to search for essential knowledge from external knowledge bases by harnessing their own strong generalizability. Specifically, we design a simple yet effective prompt to transform retrieval into a multi-hop decision sequence, which empowers LLMs with searching knowledge ability in zero-shot manner. Additionally, KSL is able to provide complete retrieval paths and therefore increase explainability of LLMs' reasoning processes. We conduct experiments on three datasets: CommonsenseQA~\citep{talmor2018commonsenseqa}, OpenbookQA~\citep{mihaylov2018openbookqa}, and MedQA-USMLE~\citep{jin2021disease}, and found that our approach improves LLM baseline performance by a relatively large margin.
\end{abstract}

\section{Introduction}
Recently, large language models (LLMs) like ChatGPT have drawn numerous attention from researchers and practitioners due to their \emph{generalist} capabilities~\citep{qin2023chatgpt}. For instance, sufficiently large language models could perform well for different tasks in zero-shot manner, such as text summarization~\citep{yang2023textgeneration, zhang2023benchmarking}, machine translation~\citep{moslem2023translation}, and question answering~\citep{singhal2023googlemedical}. However, in some scenarios, LLMs lack domain-specific knowledge or are not able to recall facts and knowledge correctly, which causes hallucination~\citep{bang2023multitask}. Hallucination refers to models generating text that is nonsensical, or unfaithful to the provided source input~\citep{ji2023survey,koehn2017six,raunak2021curious,rohrbach2018object,vinyals2015neural,maynez2020faithfulness}.

\begin{figure}[t!]
\centering
\includegraphics[width=.7\columnwidth]{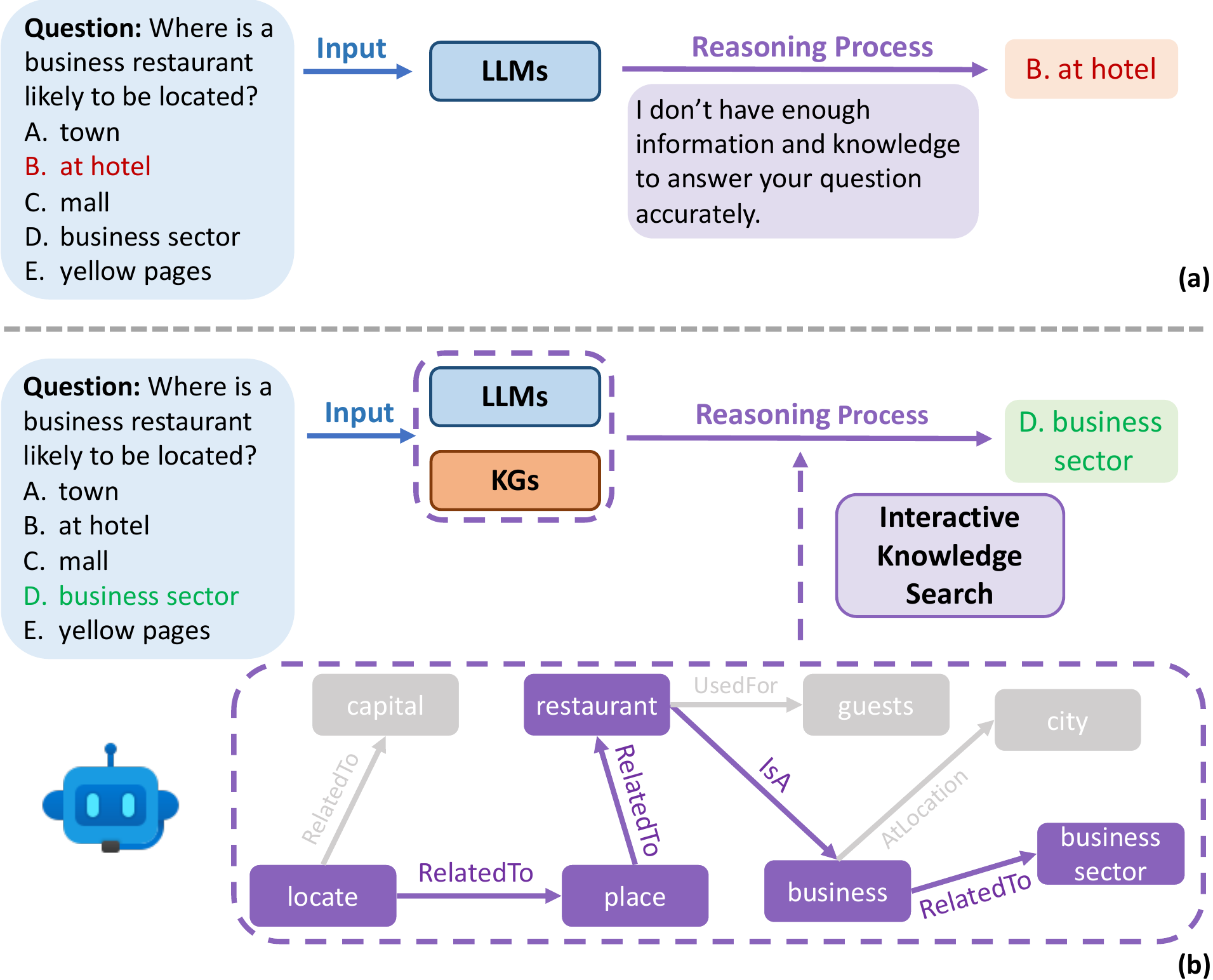} 
\caption{\textbf{Knowledge Solver.} An example comparing the vanilla LLM in (a) and zero-shot knowledge solver in (b) for question-answering tasks. Our approach helps LLMs search for necessary knowledge to perform tasks by harnessing LLMs' own generalizability. Purple represents nodes and relations in LLMs' chosen correct path.}

\label{fig1}
\end{figure}

Retrieving relevant texts from knowledge bases is a classic way to augment language models' performance like generation quality~\citep{borgeaud2022improving,lewis2020retrieval,levine2022standing,guu2020retrieval}. Besides, it can also help improve the factuality of generated texts. Typically, retrieval modules are employed to find the most relevant documents with the highest similarity scores to the query.
Then input texts and retrieved documents would be combined in a specific way fed into models. Motivated by this, some methods~\citep{ram2023context,peng2023check} utilize retrieved texts to augment LLMs. \cite{ram2023context} directly prepends retrieved documents to the input to obtain a performance gain for LLMs. ~\citep{peng2023check} designs an LLM-Augmenter to retrieve and merge evidence from external knowledge for alleviating hallucination. However, relying on similarity between embeddings would only make model learn shallow features instead of understanding semantics, which in turn hinder the model from searching truly useful knowledge. On the contrary, Knowledge Graphs (KGs) are clear, logical, and superior mediums of knowledge. Thus, effectively leveraging KGs for LLMs should benefit LLMs' performance on knowledge-required tasks.

For this reason, there is a line of work~\citep{yasunaga2022qagnn,lin2019kagnet,feng2020scalablemhgrn} using KGs to help LLMs make predictions. KagNet~\citep{lin2019kagnet} proposes a graph neural network module to model relational graphs for relational reasoning under the context of both knowledge symbolic space and language semantic space. MHGRN~\citep{feng2020scalablemhgrn} equips pretrained language models with a multi-hop relational reasoning module, which unifies path-based reasoning methods and graph neural networks. QA-GNN~\citep{yasunaga2022qagnn} learn representations over joint graphs formed by connecting QA context and KG. However, they~\citep{yasunaga2022qagnn,lin2019kagnet,feng2020scalablemhgrn} all require training additional knowledge-aware modules like graph neural networks (GNNs) on retrieved knowledge. There are two shortcomings of training additional modules: 1) would suffer from pains of retraining when encountering novel domains; 2) would become a bottleneck since LLMs’ strong abilities are not fully utilized for retrieval.

In this paper, we propose a paradigm, termed Knowledge Solver (KSL), to solve these shortcomings, which teaches LLMs themselves to search for knowledge from external knowledge bases. To be specific, we simplify the process of searching for necessary knowledge from KGs into a multi-hop decision sequence. At each step, we transform local information within KGs into text prompts (including the historical path selected by LLMs), based on which LLMs select relevant knowledge in the context to perform tasks, as shown in Figure~\ref{fig1}. The whole process is similar to humans searching over the Internet for achieving some goals. Furthermore, based on the complete paths chosen by LLMs, we can explain the whole decision-making process of LLMs. It allows for analysis when bad cases arise, a capability not present in previous black-box retrieval methods.

We evaluate our approach, Knowledge Solver (KSL), with three LLMs (GPT-3.5, LLaMA~\citep{touvron2023llama}, and LLaMA 2~\citep{Touvron2023Llama2O}) on three datasets: CommonsenseQA, OpenbookQA, and MedQA-USMLE, where reasoning with knowledge is required. KSL improves two LLM baselines' performance across these three datasets in zero-shot and finetuning settings.

Our main contributions are summarized as follows:
\begin{itemize}
\item We propose Knowledge Solver (KSL), which is the first paradigm employing LLMs to search for relevant knowledge on KGs by themselves.
\item Our proposed paradigm Knowledge Solver can boost LLMs' performance on knowledge-required tasks by a relatively large margin in zero-shot manner without additional modules and training.
\item Knowledge Solver can provide explainability for LLMs' whole reasoning processes.
\item When the computational burden is affordable, finetuning LLMs on our specially constructed dataset, with the help of KGs, can benefit LLMs further. 
\end{itemize}

\section{Related Work}
\paragraph{Large Language Models.}
Pre-trained language models (PLMs) are trained on massive datasets, which enables them to understand contexts and generate texts. Pre-trained LMs like GPT-1~\citep{radford2018gpt1}, BERT~\citep{devlin2019bert}, XLNet~\citep{yang2020xlnet},~RoBERTa \cite{liu2019roberta} and ALBERT~\citep{lan2020albert} have been widely applied to various natural language processing (NLP) tasks in recent years. For the task of question answering, models are leveraged in a large number of existing frameworks, such as ~\citep{lin2019kagnet, lvaaai2020graphbased, feng2020scalablemhgrn, yasunaga2022qagnn, zhang2022greaselm} to encode the QA contexts as statement vectors.

The current burst of development in large language models (LLMs) brings new innovation hits with the immense size and capacity. Base LLMs like T5~\citep{raffel2020exploring}, GPT-3~\citep{brown2020language}, PaLM~\citep{chowdhery2022palm}, GPT-J~\citep{gptj}, LLaMA~\citep{touvron2023llama}, GLM~\citep{glm, zeng2022glm}, BLOOM~\citep{scao2022bloom}, RWKV~\citep{peng2023rwkv}, MOSS~\citep{sun2023moss} and LLaMA 2~\citep{Touvron2023Llama2O} are trained on large datasets to capture general language patterns. Additionally, instruction fine-tuned LLMs like InstructGPT~\citep{ouyang2022instructgpt}, Flan-PaLM~\citep{chung2022flanpalmt5}, Flan-T5~\citep{chung2022flanpalmt5}, BLOOMZ~\citep{muennighoff2022bloomz}, Alpaca~\citep{alpaca} and Vicuna~\citep{vicuna2023} are designed to follow user instructions. RLHF (Reinforcement Learning from Human Feedback) LLMs, such as ChatGPT\footnote{\url{https://openai.com/blog/chatgpt/}} and GPT-4~\citep{openai2023gpt4}, incorporate reinforcement learning techniques to optimize model performance based on human feedback. However, in some scenarios, LLMs 
lack domain-specific knowledge to perform relevant tasks. Our proposed paradigm, KSL, teaches LLMs themselves to search for knowledge from external knowledge bases to help LLMs achieve goals. 

\paragraph{Knowledge Base Question Answering.}
Question answering over knowledge base (KBQA) focuses on enabling machines to answer questions using relevant knowledge retrieved from knowledge bases (KBs). Approaches in KBQA can be broadly categorized into two groups: (i) text retrieval-based methods and (ii) Knowledge Graph-based methods. Our research aligns with the second group, with an emphasis on integrating Knowledge Graphs into LLMs.

\begin{figure*}[t!]
    \centering
    \includegraphics[width=\textwidth]{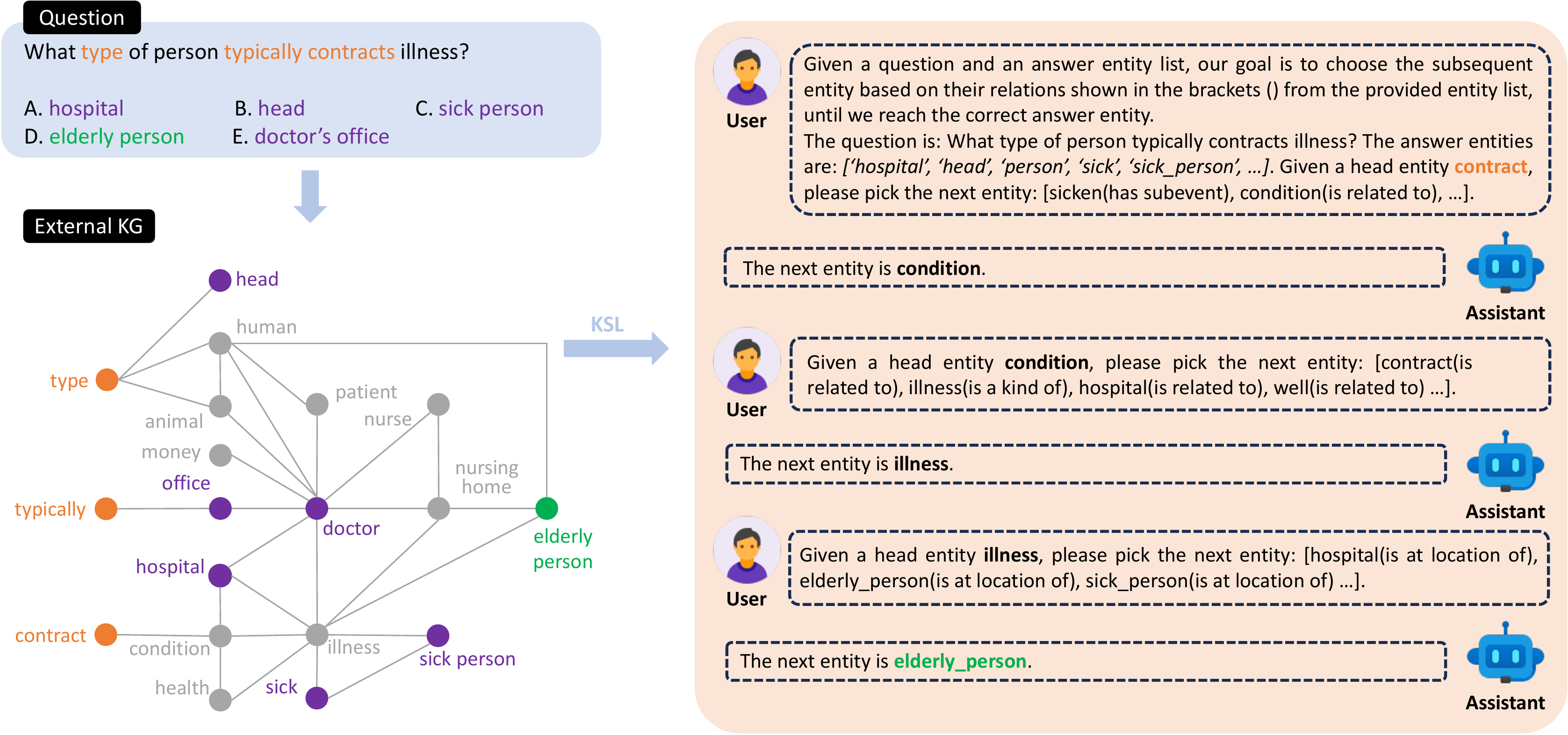} 
    \caption{\textbf{Method Overview.} For each question answer choice pair, we retrieve relevant knowledge subgraph and encode it into text prompt, which is injected into LLMs directly to help them perform knowledge-required tasks. In this question-answering scenario, LLMs interact with provided external knowledge to choose the path for answering the question correctly.}
    \label{fig2}
\end{figure*}

Text retrieval-based methods have been experimented with a wide range of NLP tasks. Generative models, augmented with retrieval capabilities in question answering, are studied (and finetuned) in~\citep{min2020ambigqa, lewis2020retrievalrag, izacard2020leveraging}. Rather than directly finetuning pretrained LMs to enhance language task performance, a growing number of researchers are moving towards lighter-weight approaches, where they freeze model parameters and augment the model with small trainable modules. Such lightweight finetuning techniques include adapter tuning~\citep{houlsby2019parameter, lin2020exploring}, prompt tuning~\citep{lester2021prompt}, prefix tuning~\citep{li2021prefix}, and more complex architectures like input-dependent
prompt tuning, frozen readers, and LM recursion as presented in ~\citep{levine2022standing}.

Knowledge Graph-based methods are also widely applied in the question answering domain. KagNet~\citep{lin2019kagnet} constructs schema graphs representing paths between question and answer entities, which are then encoded with GCN-LSTM-HPA architecture. To achieve both high accuracy and effective model scalability, Multi-hop Graph Relation Network~\citep{feng2020scalablemhgrn} combines path-based reasoning interpretability with GNN scalability, adding a structured relational attention mechanism. Distinctly, QA-GNN~\citep{yasunaga2022qagnn} links QA context vectors to topic entities in the schema graph. DRAGON~\citep{yasunaga2022dragon} proposes a self-supervised model for bidirectional text and KG integration, while GreaseLM~\citep{zhang2022greaselm} fuses PLMs and GNN representations through layered modality interactions. Unlike prior works training additional modules like GNNs, our method KSL encourages LLMs to search for essential knowledge from external knowledge base by themselves. 

\section{Problem Definition}
Our paper aims to help LLMs perform better on knowledge-required tasks when they lack domain-specific knowledge. We choose question answering as the evaluated knowledge-required task. To mitigate the issue of lacking knowledge, we inspire LLMs to interact with provided external knowledge and spontaneously identify the appropriate pathway to derive the correct answer. Following prior work~\citep{yasunaga2022qagnn}, we define the Knowledge Graph as a multi-relational graph $\mathcal{G}$ = ($\mathcal{V}$, $\mathcal{E}$). Here $\mathcal{V}$ is the set of entity nodes in the KG; $\mathcal{E} \in \mathcal{V} \times \mathcal{R} \times \mathcal{V} $ is the set of edges that connect nodes in $\mathcal{V}$, where R represents a set of relation types.

Given question answer choices pair $[q, A]$, we link entities mentioned in the question and answer choices to the given KG $\mathcal{G}$, following prior work~\citep{feng2020scalablemhgrn}. We denote all question entities as $\mathcal{V}_{q} \in \mathcal{V}$, and answer entities as $\mathcal{V}_{a} \in \mathcal{V}$. Then we retrieve subgraph $\mathcal{G}_{sub} = (\mathcal{V}_{sub}^{q, a}, \mathcal{E}_{sub}^{q, a})$ from KG $\mathcal{G}$. $\mathcal{G}_{sub}$ contains all nodes on the k-hop paths between nodes in $\mathcal{V}_{q}$ and $\mathcal{V}_{a}$.

\section{Method}
\begin{figure}[t!]
\centering
\footnotesize
\resizebox{.85\linewidth}{!}{
\begin{minipage}{\linewidth}
    \begin{algorithm}[H]
    \footnotesize
    \begin{algorithmic}[1]
    \Require
    Question entities $\mathcal{V}_{q}=\{v_{q1}, v_{q2}, \cdots, v_{qn}\}$; corresponding answer entities  $\mathcal{V}_{a} = \{v_{a1}, v_{a2}, \cdots, v_{an}\}$. 
    \Function{rel\_extr}{$v_{h}$, $\mathcal{G}_{sub}$}
    \State
    tail\_relation\_list = []
    \For{each tail entity $v_{ti}$ of $v_{h}$ in $\mathcal{G}_{sub}$}
    \State
    relation $r_{hti} = \mathcal{G}_{sub}(v_{h},v_{ti})$
    \State
    tail\_relation\_list.append(($v_{ti},r_{hti}$))
    \EndFor
    \State
    return tail\_relation\_list
    \EndFunction
    \State
    retrieve subgraph $\mathcal{G}_{sub}$ given $\mathcal{V}_{q}$ and $\mathcal{V}_{a}$
    \State
    $v_{q}$ is randomly selected from $\mathcal{V}_{q}$ as $v_{h1}$
    \State
    round = 0
    \For{each head entity $v_{hi}$}
    \If{$v_{hi} \in \mathcal{V}_{a}$ }
    \State
    break
    \EndIf
    \If{round == round\_maximum}
    \State
    break
    \EndIf
    \State
    tail\_relation\_list = REL\_EXTR($v_{hi}$, $\mathcal{G}_{sub}$)
    \State 
    $v_{h(i+1)}$ = LLM(tail\_relation\_list)
    \State
    round += 1
    \EndFor
    \State
    \Return $v_{hi}$
    \end{algorithmic}
    \caption{Knowledge Solver Zero-Shot Reasoning.}
    \label{alg:inference}
    \end{algorithm}
\end{minipage}
}
\end{figure}
As shown in Figure~\ref{fig2}, our method KSL first retrieves relevant subgraph $\mathcal{G}_{sub}$ from KG for given question answer choices pair $[q, A]$. Then we encode $\mathcal{G}_{sub}$ into text prompt $T_{K}$ to inject knowledge into LLMs, which would initialize dialogue-like inference to encourage LLMs to search necessary knowledge by utilizing their own abilities and guide themselves to achieve final goals.

\subsection{Knowledge Solver Zero-Shot Reasoning}\label{kszs}

In order to help models perform tasks that require domain-specific knowledge, like question answering, we inject external knowledge into LLMs. For each retrieved subgraph $\mathcal{G}_{sub}$, we transform it into text prompt $T_{K}$ fed into LLMs, and utilize LLMs' strong generalizability to incentivize them to search for necessary information by themselves. 

Given the question $q$ and the set of answer choices $A = [a_{1}, ..., a_{N}]$, where N is the total number of answer choices, we retrieve $\mathcal{G}_{sub}$ and view it as external knowledge. The $\mathcal{G}_{sub}$ contains all question entities $\mathcal{V}_{q}$, all answer choice entities $\mathcal{V}_{a}$, intermediate entities, and corresponding relations $\mathcal{R}$ between entities. To initialize the reasoning process of LLMs for question answering, we first randomly select a question entity $v_{q} \in \mathcal{V}_{q}$ for LLMs, and then encourage LLMs to choose a path based on their own judgment until they finally reach one of the answer entities $v_{a} \in \mathcal{V}_{a}$. Concretely, we can break down the reasoning process of LLMs for question answering into several rounds like CoT~\citep{wei2022chain} (the total number of rounds depends on LLMs' own judgment. In practice, we set the limit of rounds to $N_{r}$). For each question and answer choices pair $[q, A]$, the chain of rounds would form an explicit reasoning path, which not only augments LLMs with domain-specific external knowledge, but also increases LLMs' explainability. 

During each round, we put the current head entity $v_{h}$ and all linked head entities $\mathcal{V}_{t} = [v_{t1}, ..., v_{tN}]$ and their corresponding relations $\mathcal{R}_{ht} = [r_{ht1}, ..., r_{htN}]$ in the text prompt to inform LLMs of the existence of external knowledge. LLMs will pick the most likely tail entity as the head entity for the next round, based on the prior knowledge implicitly stored in their parameters and explicit external knowledge in the form of text prompts, like relations, for question answering. Then, this entity selection process will repeat until one of the answer entities $v_{a}$ is chosen. Ultimately, we find the LLMs' selected answer choice based on the mapping between answer entity $v_{a}$ and answer choice $a$. The whole reasoning process is purely done by text generation instead of classification over predefined entities since in many scenarios, we are not able to access the logits of LLMs. For each round, the input text prompt also includes the whole history of entity selection, similar to dialogue. The overall reasoning process is also illustrated in Algorithm~\ref{alg:inference}.

\subsection{Knowledge Solver Finetuning}
\begin{figure}[t!]
\centering
\footnotesize
\resizebox{.85\linewidth}{!}{
\begin{minipage}{\linewidth}
    \begin{algorithm}[H]
    \footnotesize
    \begin{algorithmic}[1]
    \Require
    A sequence of all question answer choices pairs $Q = \{[q_1, A_1], \cdots, [q_N, A_N]\}$; structured knowledge source (Knowledge Graph) $\mathcal{G}$; encoder $\mathcal{E}$ to transform $\mathcal{G}_{sub}$ into text prompt $T_K$ 
    
    \State total\_paths = []
    \For{each $[q_j, A_j]$ in $Q$}
    \State extract question and answer choices entities $\mathcal{V}_{q}$ and $\mathcal{V}_{a}$
    \State retrieve subgraph $\mathcal{G}_{sub}$ from $\mathcal{G}$
    \For{each question entity $v_{qi}\in \mathcal{V}_{q}$}
    \State randomly select correct answer choice entity $v_{ca}$
    \State path = find\_shortest\_path($\mathcal{G}_{sub}$, source=$v_{qi}$, target=$v_{ca}$)
    \State total\_paths.append(path)
    \State remove all nodes on the path except for $v_{ca}$ from ${G}_{sub}$
    \EndFor
    \EndFor
    \State training\_data = []
    \For{each path $p_i$ in total\_paths }
    \State
    hist = []
    \For{each node $n_j$ in $p_i$ except for the last node}
    \State
    instance = \{\}
    \State
    instance[``instruction"] = instruction
    \State head entity $v_{hj}$ = $n_j$
    \State tail\_entity $v_{tj}$ = entity\_extract($\mathcal{G}_{sub}$, $v_{hj}$)
    \State relation$r_{htj}$ = relation\_extract($\mathcal{G}_{sub}$, $v_{hj}$, $v_{tj}$)
    \State
    instance[``input"] = $\mathcal{E}$($v_{hj}$, hist)
    \State
    instance[``output"] = $\mathcal{E}$($v_{tj}$)
    \State
    training\_data.append(instance)
    \State
    hist.append([instance[``input"], instance[``output"]])
    \EndFor
    \EndFor
    \State
    \Return training\_data

    \caption{Generating Training Instruction Dataset.}
    \label{alg:generate_dateset}
    \end{algorithmic}
    \end{algorithm}
    
\end{minipage}
}
\end{figure}

When LLMs are accessible, we can finetune them on external knowledge to transform this knowledge into LLMs' parameters. Following Alpaca~\citep{alpaca}, we leverage instruction tuning~\citep{wei2021finetuned} to finetune LLMs.

\begin{figure}[t!]
    \centering
    \includegraphics[width=.5\linewidth]{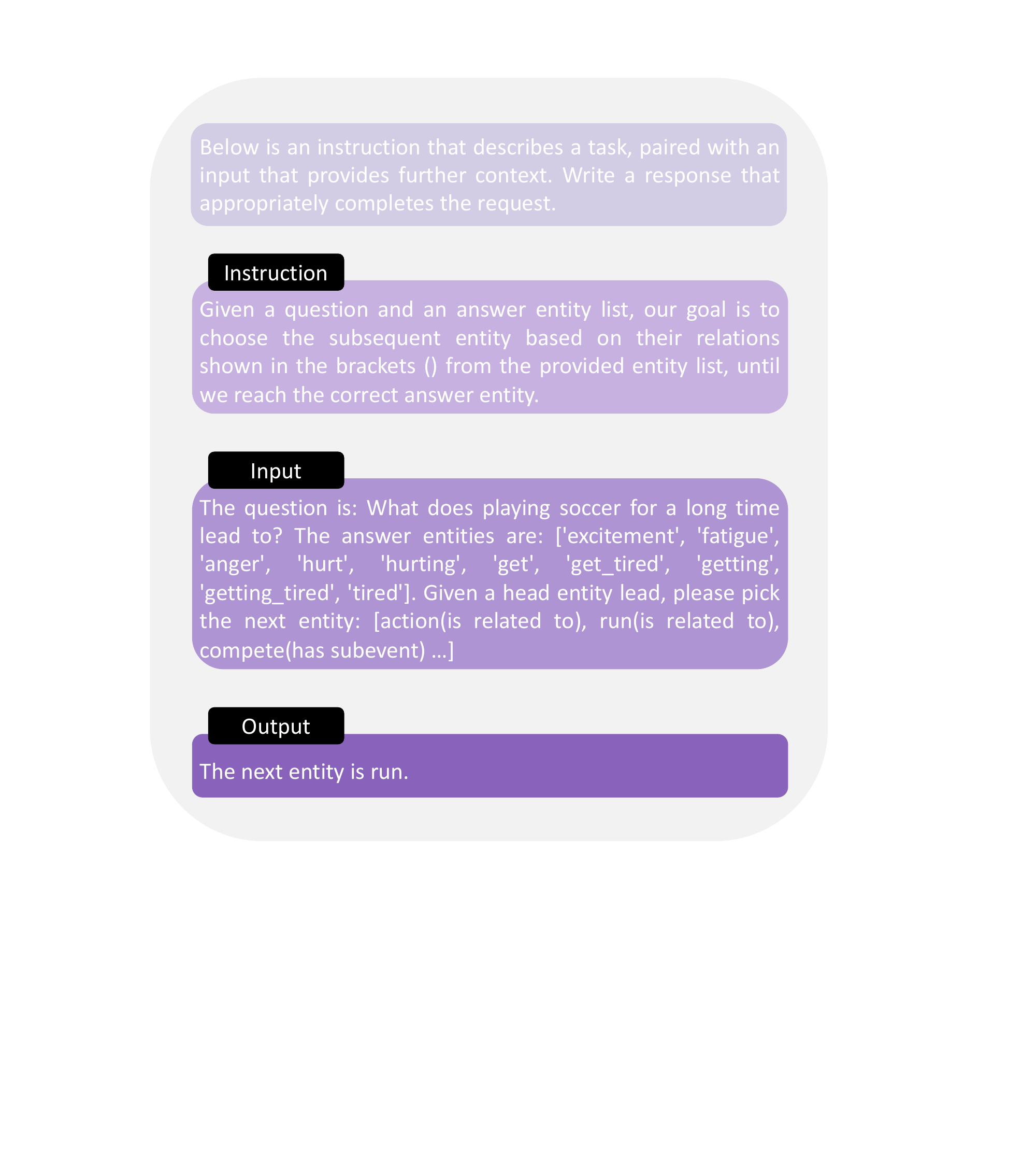} 
    \caption{\textbf{Training example.} Instance in our constructed instruction tuning dataset.}
    \label{fig3}
\end{figure}

To be specific, we use a similar template in Alpaca~\citep{alpaca}. Different from general instruction tuning~\citep{wei2021finetuned}, where LLMs are stimulated to follow users' instructions in zero-shot manner, our main goal is to encourage LLMs to learn domain-specific knowledge. Thus, we fix instructions, which are used to inform LLMs to select the correct path, across all instances (in reality, the instructions can be modified according to domain-specific knowledge). The input and response formats are the same as we stated in Knowledge Solver Zero-Shot Reasoning, where we transform each retrieved subgraph $\mathcal{G}_{sub}$ into multiple input-response pairs starting from question entity $v_{q}$ to answer entity $v_{a}$ in the correct answer choice. Each input contains entity selection history like the dialogue between the user and LLMs, the current head entity, all connected tail entities, and corresponding relations. The response includes the next tail entity of the correct path. Concretely, for each question and answer choices pair $[q, A]$, we iterate over all question entities $v_{q} \in \mathcal{V}_{q}$ while keeping all extracted paths separated. The whole process of constructing instruction-tuning dataset is also illustrated in Algorithm~\ref{alg:generate_dateset}. The example of our instruction tuning dataset can be seen in Figure~\ref{fig3}. We utilize LoRA~\citep{hu2021lora} to tune LLMs since it can help greatly reduce GPU memory burden. 

For inference, finetuned KSL uses the same way as zero-shot Knowledge Solver. For each question and answer choices pair $[q, A]$, we randomly select a question entity $v_{q}$ to initialize the reasoning process. We leave averaging results of all question entities for future research.
\section{Experiment}

\subsection{Datasets}
We evaluate our approach Knowledge Solver on three question-answering datasets: CommonsenseQA~\citep{talmor2018commonsenseqa}, OpenbookQA~\citep{mihaylov2018openbookqa}, and MedQA-USMLE~\citep{jin2021disease}. 

\paragraph{CommonsenseQA} is a question-answering dataset for commonsense reasoning, comprising a total of 12102 questions. The methodology for question generation involves sampling three target concepts related to a source concept from ConceptNet~\citep{speer2017conceptnet}. Each question has five choices. Three of these are authored by crowd workers based on the target concepts, with an additional two serving as distractors.
CommonsenseQA serves as one of the most common benchmark datasets for KGQA, as shown in~\citep{lin2019kagnet, lvaaai2020graphbased, feng2020scalablemhgrn, yasunaga2022qagnn, yasunaga2022dragon}. Our paper preprocesses data with the original data splits in KagNet~\citep{lin2019kagnet}.

\paragraph{OpenbookQA} contains approximately 6000 multiple-choice questions and an open book of over 1000 elementary-level science facts. The question-answering process requires a combination of scientific facts, commonsense knowledge, and multi-hoop reasoning abilities. Our paper follows the original data splits~\citep{mihaylov2018openbookqa}.

\paragraph{MedQA} is a multilingual dataset designed for solving real-world medical problems. All questions and answers are gathered from professional medical board exams. In our paper, we focus on the USMLE subset, where data is from the National Medical Board Examination in the USA, and follow the original data splits~\citep{jin2021disease}.
\begin{table}[t!]
\centering
\resizebox{0.6\linewidth}{!}{\begin{tabular}{llll}
\toprule
Models     & CSQA & OBQA & MedQA \\
\midrule
GPT-3.5 (zero-shot) &   72.9      &   74.8      &   55.8      \\
GPT-3.5 + KSL (zero-shot)  &  \textbf{79.6} (+9.19\%)       &    \textbf{81.6} (+9.09\%)      &  
\textbf{58.4} (+4.66\%)        \\
\midrule
LLaMA-7B (zero-shot)    &  20.5       &   26.8       &  22.7       \\
LLaMA-7B +  KSL (zero-shot)   & \textbf{28.4} (+38.54\%)       &    \textbf{34.0} (+26.87\%)      &  \textbf{23.6} (+3.96\%)       \\
LLaMA2-7B (zero-shot)    &  19.7      &    25.6       &    25.1    \\
LLaMA2-7B + KSL (zero-shot)   &   \textbf{26.3} (+33.50\%)     &      \textbf{32.2} (+25.78\%)  &   \textbf{25.8} (+2.79\%)      \\
\midrule
LLaMA-7B (finetuned)    &   38.0     &  29.8      &  25.0   \\ 
LLaMA-7B +  KSL (finetuned)   &   \textbf{47.4} (+24.74\%)        &    \textbf{45.8} (+53.69\%)       &      \textbf{25.7} (+2.80\%)   \\
\bottomrule \\
\end{tabular}}
\caption{\textbf{Performance Evaluation.} We report the accuracy of LLM baselines and (zero-shot and finetuned) KSL on three datasets: CommonsenseQA, OpenBookQA, and MedQA-USMLE.}
\label{result}
\end{table}
\subsection{Knowledge Graphs}

CoceptNet~\citep{speer2017conceptnet} is used for CommonsenseQA and OpenbookQA. It links words and phrases from common human language via labeled relationships. We adopt the relation setups from MHGRN~\citep{feng2020scalablemhgrn}, which include a total of 34 multi-directional relation types. The paths between all topic entities mentioned in the question-answer pair are founded and grounded as the subgraphs. 

In the context of the USMLE dataset of MedQA, we incorporate the Knowledge Graph constructed in QA-GNN~\citep{yasunaga2022qagnn}, which contains biomedical vocabularies from Unified Medical Language System (UMLS)~\citep{bodenreider2004umls} and DrugBank~\citep{wishart2018drugbank}.

Given each question and answer choices pair $[q, A]$, we retrieve subgraph $\mathcal{G}_{sub}$ from structured Knowledge Graph $\mathcal{G}$ following the preprocessing step described in MHGRN~\citep{feng2020scalablemhgrn}, with hop size $k=2$.

\subsection{Implementation \& training details}

\paragraph{Zero-shot.} We mainly use three LLMs (GPT-3.5, LLaMA-7B~\citep{touvron2023llama}, and LLaMA 2-7B~\citep{Touvron2023Llama2O}) as baselines. For GPT-3.5, we call OpenAI API to use gpt-3.5-turbo-16k. The limit of the total number of rounds $N_{r}$ is set to 5 during inference. 

\begin{figure*}[t!]
    \centering
    \includegraphics[width=0.85\linewidth]{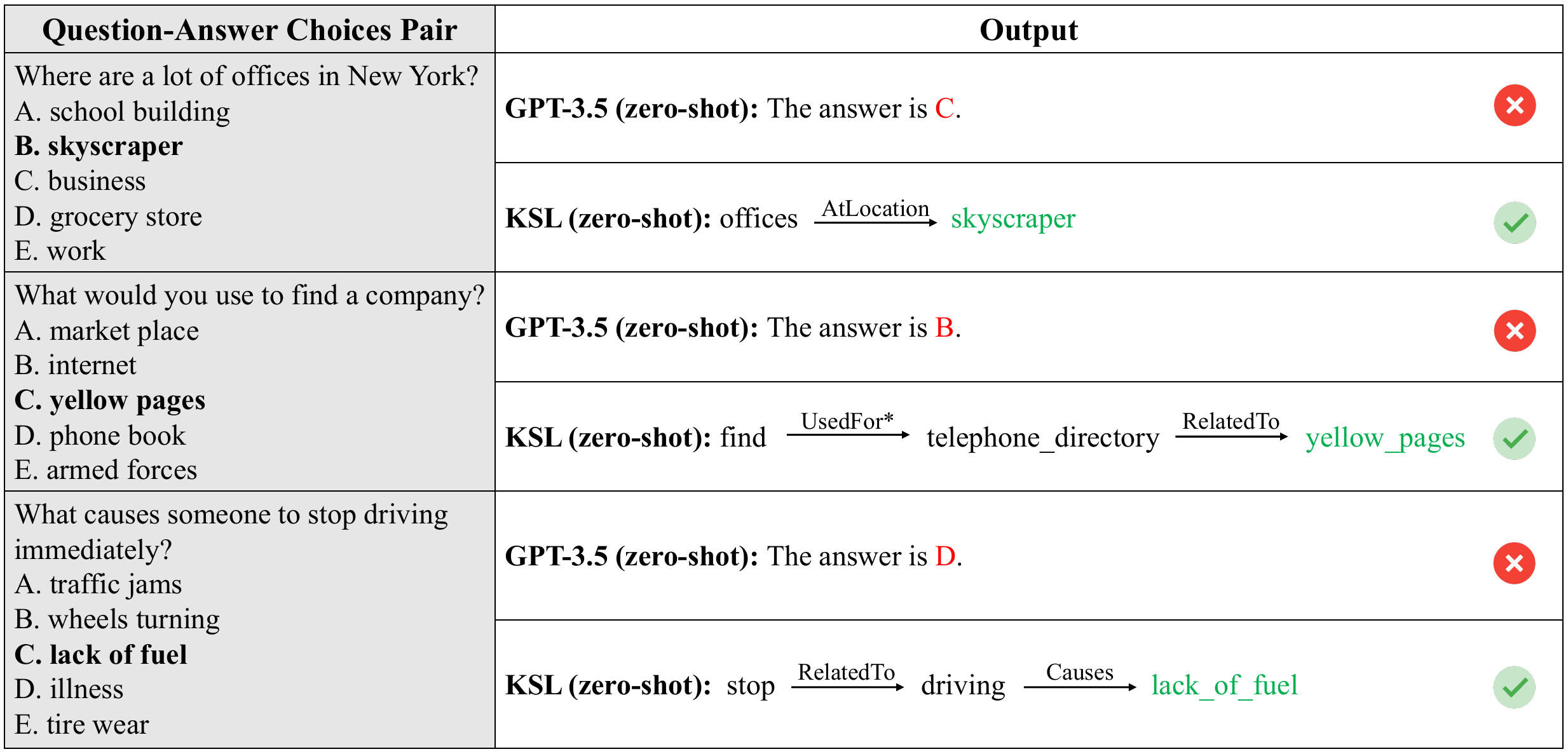} 
    \caption{\textbf{Qualitative Results of KSL (GPT-3.5).} Generated responses on some examples of GPT-3.5 and zero-shot KSL (GPT-3.5). The bold choice represents the correct answer. An asterisk (*) denotes a reversed relation.}
    \label{fig5}
\end{figure*}

\paragraph{Finetuning.} We use LoRA~\citep{hu2021lora} to finetune LLaMA-7B~\citep{touvron2023llama} on 8 NVIDIA A40 GPUs, each has 48 GB memory. For CommonsenseQA~\citep{talmor2018commonsenseqa}, the training set contains 114,552 instances and the development set consists of 14,391 instances. For OpenbookQA~\citep{mihaylov2018openbookqa}, the training set includes 57,458 instances and the development set contains 5814 examples. For MedQA-USMLE~\citep{jin2021disease}, there are 13,561 instances in the training set and 1677 instances in the development set. The global batch size is 128 and learning rate is set to 3e-4. We set the rank $r$ in LoRA~\citep{hu2021lora} to 16 and $\alpha$ to 16. The dropout probability~\citep{srivastava2014dropout} is 0.05. We finetune query, key, value, and output projection matrices $W_{q}, W_{k}, W_{v}, W_{o}$ in self-attention modules of transformers~\citep{vaswani2017attention}. The maximum of input sequence length is 1152. The total number of finetuning epochs for CommonsenseQA~\citep{talmor2018commonsenseqa} and OpenbookQA~\citep{mihaylov2018openbookqa} is 3, and for MedQA-USMLE~\citep{jin2021disease} is 5. We use checkpoints with the lowest validation loss for final inference on test sets. 

\paragraph{Evaluation metric.} For three question-answering datasets: CommonsenseQA~\citep{talmor2018commonsenseqa}, OpenbookQA~\citep{mihaylov2018openbookqa}, and MedQA-USMLE~\citep{jin2021disease}, we use accuracy as evaluation metric following prior work~\citep{yasunaga2022qagnn}. However, we only perform text generation instead of classification over the predefined set, it is hard to use the traditional way for calculating accuracy. Instead, we call OpenAI API and input hand-crafted prompts (see details in supplementary) to GPT-4~\citep{openai2023gpt} to judge whether LLMs' generation matches ground truth. In the end, we use the score from GPT-4~\citep{openai2023gpt} for calculating accuracy (0 represents the LLMs' output is totally irrelevant while 1 means that LLMs' generated answer correctly matches the ground truth).

\subsection{Result}

\begin{figure*}[htbp]
    \centering
    \includegraphics[width=0.85\linewidth]{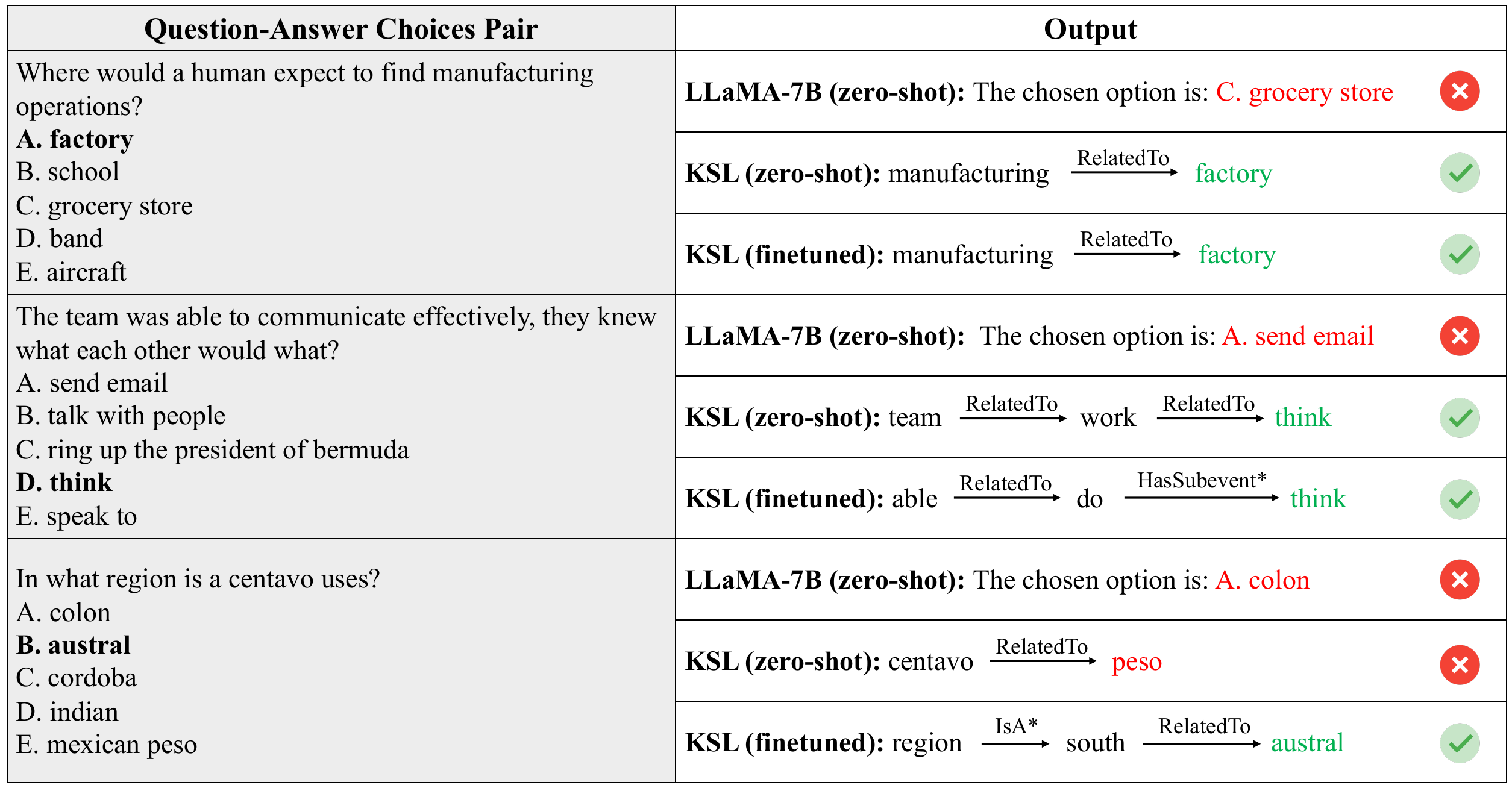} 
    \caption{\textbf{Qualitative Results of KSL (LLaMA-7B).} Generated responses on some examples of LLaMA-7B and zero-shot/finetuned KSL (LLaMA-7B). The bold choice represents the correct answer. An asterisk (*) denotes a reversed relation.}
    \label{fig6}
\end{figure*}

\paragraph{Knowledge Solver zero-shot reasoning.} As shown in Table~\ref{result}, our Knowledge Solver (KSL) can boost LLM baselines (GPT-3.5, LLaMA-7B~\citep{touvron2023llama}, and LLaMA 2-7B~\citep{Touvron2023Llama2O}) by a relatively large margin, indicating that: 1) our approach can benefit model in performing knowledge required tasks; 2) LLMs possess certain abilities to search necessary information by themselves when external knowledge is provided. Training an adapter for each scenario where domain-specific knowledge is required would cast large computational and time costs. In contrast, our zero-shot Knowledge Solver can harness LLMs' own emergent ability to perform domain knowledge-required tasks by only providing external knowledge. This teaches LLMs to interact with external knowledge to achieve final goals. 

\paragraph{Knowledge Solver finetuning.}
Unlike training separate adapters like GNNs, our approach can also finetune LLMs on provided external knowledge to inject knowledge into LLMs' parameters. As shown in Table~\ref{result}, finetuned KSL (LLaMA-7B) can improve performance further and surpass finetuned LLaMA-7B (see finetuning detials in supp.) on three datasets. 
This suggests that our method can effectively help LLMs memorize knowledge to perform domain-specific knowledge-required tasks when the computational burden is affordable. Interestingly, the improvement on MedQA-USMLE~\citep{jin2021disease} is not as substantial as on CommonsenseQA~\citep{talmor2018commonsenseqa} and OpenBookQA~\citep{mihaylov2018openbookqa}. The problem might be due to the fact that Knowledge Graph~\citep{yasunaga2022qagnn} is not large enough, where for many question and answer choices pairs, it is difficult to retrieve complete subgraphs. In many cases, several answer entities are not included in subgraphs or there is no path from question entities to answer entities. 

\subsection{Qualitative result}
We show some qualitative results in Figure~\ref{fig5} and Figure~\ref{fig6}. It shows that our zero-shot KSL can help LLMs perform knowledge-required tasks without any additional training. Provided with external knowledge, LLMs can look up necessary knowledge to achieve final goals by themselves. Our approach can help LLMs correct their mistakes when they lack relevant domain-specific knowledge. For example, vanilla LLaMA-7B~\citep{touvron2023llama} doesn't know where the manufacturing operations can be found while zero-shot KSL (LLaMA-7B) can correctly answer the question. Finetuned KSL (LLaMA-7B) can further improve LLMs' ability to solve knowledge-required tasks like answering the question ``in what region is a centavo uses?". These demonstrate the effectiveness of KSL.
\subsection{Ablation study}
\begin{figure}[t!]
    \centering
    \includegraphics[width=0.65\linewidth]{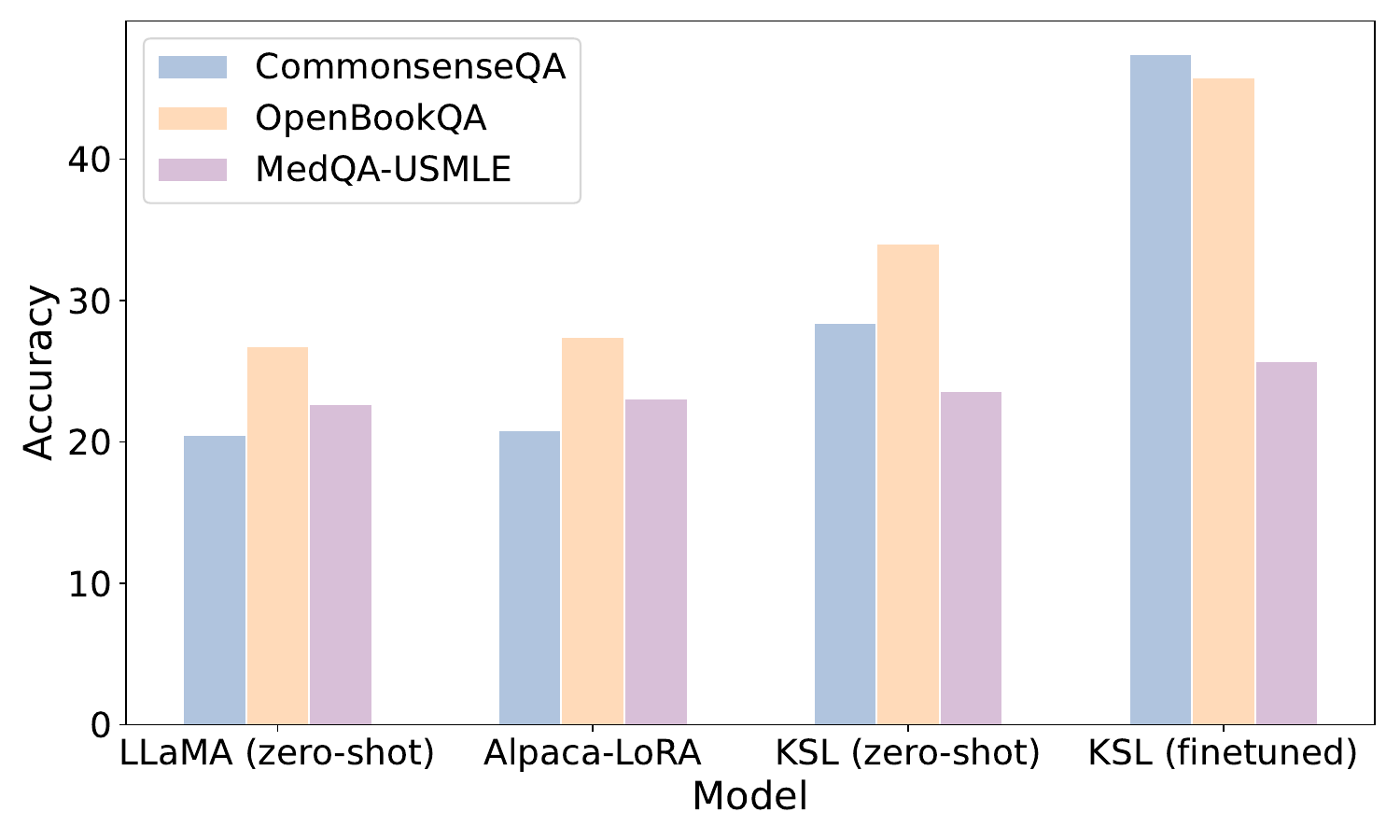} 
    \caption{\textbf{Ablation Experiments on Finetuned KSL (LLaMA-7B).} We compare our KSL with LLaMA and Alpaca-LoRA.}
    \label{fig4}
\end{figure}
As shown in Table~\ref{result}, finetuned KSL (LLaMA-7B) can improve performance substantially. In order to investigate whether this boost mainly comes from instruction tuning itself or our specially constructed knowledge datasets, we also evaluate Alpaca-LoRA~\citep{alpaca} on CommonsenseQA~\citep{talmor2018commonsenseqa}, OpenBookQA~\citep{mihaylov2018openbookqa}, and MedQA-USMLE~\citep{jin2021disease} by using the same inference method mentioned as vanilla LLaMA-7B~\citep{touvron2023llama}. It's worth noting that Alpaca-LoRA's maximum sequence length is 512, while for our interactive inference method, the input sequence length is generally longer than 512. As shown in Figure~\ref{fig4}, Alpaca-Lora, which uses the same technique LoRA~\citep{hu2021lora} tuning LLaMA-7B~\citep{touvron2023llama}, works on par with vanilla LLaMA-7B~\citep{touvron2023llama}, suggesting that our specially designed knowledge dataset is the main source benefiting LLMs on performing knowledge required tasks. Alpaca-LoRA~\citep{alpaca} underperforms our zero-shot KSL (LLaMA-7B). It indicates that encouraging LLMs to search for relevant knowledge by harnessing their own abilities is an effective and efficient way to help model on knowledge-required tasks. 

\section{Conclusion}
In this paper, we propose Knowledge Solver (KSL), which can help LLMs perform better on domain-specific knowledge-required tasks in zero-shot and finetuning manner. Provided with external knowledge, LLMs can harness their own ability to search for necessary knowledge and information to perform relevant tasks without additional training or modules. Our interactive inference method can not only explicitly inject knowledge into LLMs but also guide LLMs to solve tasks. We also demonstrate that performance improvement majorly comes from our specially designed inference method (for zero-shot) and task (for finetuning) instead of instruction tuning. Currently, the initial question entity for our interactive inference method is randomly chosen. We leave how to choose the first entity to initialize performing tasks for further research.

\bibliography{custom}
\bibliographystyle{apalike}

\end{document}